\documentclass[letterpaper]{article}

\usepackage[T1]{fontenc}

\usepackage{geometry}
\usepackage{setspace}

\usepackage[style = chem-acs, articletitle = true, chaptertitle = true, pageranges = true]{biblatex}
\usepackage{xpatch}
\xpatchbibdriver{inproceedings}
  {\usebibmacro{byauthor}\setunit{\addspace}}
  {\usebibmacro{byauthor}\setunit{\labelnamepunct}\usebibmacro{title}%
   \setunit{\addperiod\space}}
  {}
  {\PackageWarning{biblatex}{Could not patch the inproceedings driver}}
\xpatchbibdriver{online}
  {\printunit{\addspace}}
  {\printunit{\addperiod\space}}
  {}
  {\PackageWarning{biblatex}{Could not patch the online driver}}
\renewcommand*{\labelnamepunct}{\addperiod\space}
\usepackage{graphicx}
\usepackage{float}
\newfloat{scheme}{htbp}{los}
\floatname{scheme}{Scheme}
\floatname{chart}{Chart}
\newfloat{graph}{htbp}{loh}
\usepackage{subcaption}
\graphicspath{{figures/}}

\usepackage{amsmath}
\usepackage{amssymb}
\usepackage{url}
\usepackage[symbol]{footmisc}
\usepackage[hidelinks]{hyperref}

\newcommand*\blfootnote[1]{%
  \begingroup
  \renewcommand\thefootnote{}\footnote{#1}%
  \addtocounter{footnote}{-1}%
  \endgroup
}

\usepackage{authblk}
\author[1,$\dagger$,*]{Mateusz Praski}
\author[1,$\dagger$]{Jakub Adamczyk}
\author[1]{Wojciech Czech}
\affil[1]{Faculty of Computer Science, AGH University of Krakow, Poland}

\title{Benchmarking Pretrained Molecular Embedding Models for Molecular Representation Learning}
\date{*Email: praski@agh.edu.pl}

\begin{document}

\maketitle
\blfootnote{$^{\dagger}$ Equal contribution.}

\begin{abstract}
Pretrained molecular embedding models spanning graph neural networks, text transformers, and hybrid architectures have been widely used in drug discovery, yet no large-scale, rigorous comparison of their representations existed. We evaluated 28 models across 25 classification datasets under a controlled framework using hierarchical Bayesian Bradley-Terry statistical testing. All neural models except CheMeleon failed to outperform the classical ECFP count fingerprint; CheMeleon was the only model rated as better. However, further testing on external validation datasets, based on scaffold cross-validation and a Bayesian hierarchical correlated t-test, revealed no definite advantage of either embedding method. As a practical takeaway, we recommend evaluating both ECFP fingerprints and CheMeleon embeddings in novel QSAR/QSPR studies. Overall, in the context of all neural networks evaluated, these results also challenge the perceived progress in molecular representation learning and highlight the need for rigorous evaluation standards in the field.
\end{abstract}

\section*{Keywords}

molecular property prediction, molecular embeddings, machine learning, chemoinformatics, molecular fingerprints, representation learning

\section{Introduction}

The translation of molecular structures into information-rich numerical representations is a cornerstone of modern chemoinformatics and a critical step for harnessing machine learning in drug discovery and materials science. Molecular representation learning has emerged as a powerful paradigm, moving beyond handcrafted descriptors to automatically learn salient features from molecular data. Recently, the field has seen a surge in the development of pretrained molecular embedding models. This trend mirrors the transformative impact of representation learning in other scientific domains, where models are pretrained on vast unlabeled datasets to produce powerful general-purpose embeddings. Notable successes include models based on self-supervised learning (SSL), such as DINO \cite{DINO} and DINOv2 \cite{dinov2} in computer vision, Sentence Transformers \cite{SentenceTransformers} and TSDAE \cite{TSDAE} for dense text representations in NLP, and multimodal architectures such as CLIP \cite{CLIP} and BLIP-2 \cite{blip2} that connect images and text. Inspired by these advances, molecular models are similarly trained on massive chemical databases to generate universal embeddings.

This study focuses specifically on evaluating these static embeddings rather than the alternative approach of task-specific fine-tuning. The rationale is threefold: first, to probe the fundamental knowledge encoded during pretraining and assess the intrinsic generalization capabilities of the learned representations themselves; second, to evaluate their utility in unsupervised applications such as molecular similarity searching and clustering, where embeddings are used directly; and third, to address the challenge of low-data learning, a common scenario in chemistry where fine-tuning complex models could easily lead to severe overfitting.

The proliferation of pretrained models, each with unique architectures and pretraining objectives, has created a pressing need for their systematic and rigorous evaluation. The absence of standardized and comprehensive benchmarking of the embeddings themselves hinders the ability to make informed decisions about model selection and impedes further progress in the field. To address this critical gap, this work presents a comprehensive benchmarking study of state-of-the-art pretrained molecular embedding models. We evaluated their performance in a wide range of molecular representation learning tasks, providing a clear and objective comparison of their capabilities.

Our findings challenge the prevailing view of progress in this domain. We reveal that despite their sophistication, modern pretrained models struggle to outperform traditional, much simpler methods. Specifically, we find that molecular fingerprints often remain the top-performing representations. Furthermore, embeddings derived from Graph Neural Networks (GNNs) generally exhibit poor performance across the tested benchmarks. Although pretrained models that incorporate a strong chemical inductive bias perform acceptably, they do not demonstrate a definitive advantage. Only the CheMeleon model \cite{Chemeleon} achieved competitive performance with ECFP fingerprints under stringent statistical testing, warranting a practical recommendation to evaluate both in new QSAR/QSPR projects. These results suggest that significant progress is still required to unlock the full potential of deep learning for universal molecular representation.

\section{Molecular Embedding Approaches}

Molecular representation learning has long been a central focus in chemoinformatics and molecular machine learning, encompassing a wide range of techniques and approaches. Classical methods rely on deterministic feature extraction techniques, exemplified by molecular fingerprints. Subsequent advances in neural networks for molecular data have enabled transfer learning from large pretraining datasets through embedding models. These approaches can be categorized according to their input modality, architecture, and pretraining objectives.

Molecular graphs, based on atom and bond structures, provide a natural representation of chemical compounds. In most cases, models use the topological (2D) graph, while ignoring the spatial (3D) conformation, as 3D structures are expensive and often difficult (or even impossible) to compute. However, the 2D graph retains much of the practically useful information. Architectures that operate on this representation include Graph Neural Networks (GNNs) and graph transformers.

Compounds can also be serialized as SMILES or SELFIES strings, which are the standard formats for molecular datasets. Many models inspired by natural language processing (NLP) operate directly on these textual representations, primarily using transformer-based architectures.

Finally, there are hybrid models that utilize multimodal representations or pretraining objectives, for example, incorporating the inductive biases of graph-based representations into more scalable and easily trainable text-based models.

In the following sections, we describe a range of molecular embedding models that encompass these different approaches, all of which have been implemented for benchmarking in this study. Model selection was based on the availability of code and pretrained weights, as well as the ability to successfully run the code. We omit specific details, such as model variants or pretraining dataset curation, and refer interested readers to the original publications for further information.

\subsection{Molecular fingerprints}

Molecular fingerprints are feature extraction methods based on identifying small subgraphs within a molecule and detecting their presence or counting their occurrences, yielding binary and count variants, respectively. They can be broadly classified into substructural and hashed types \cite{scikit-fingerprints}. Substructural fingerprints detect predefined patterns, such as functional groups or ring systems, that are typically determined by expert chemists. In contrast, hashed fingerprints define general shapes of extracted subgraphs, convert them into numerical identifiers, and hash them using a modulo function into a fixed-length output vector. Common examples include circular neighborhoods in Extended Connectivity FingerPrint (ECFP) \cite{fp_ecfp}, paths of four atoms in the Topological Torsion (TT) fingerprint \cite{fp_topological_torsion}, and the shortest paths between atom pairs in the Atom Pair (AP) fingerprint \cite{fp_atom_pair}.

Although not task-adaptive like neural networks, hashed fingerprints remain widely used in chemoinformatics and molecular machine learning due to their flexibility, computational efficiency, and consistently strong performance. In many cases, they continue to outperform more complex approaches, such as GNNs \cite{fingerprints_vs_gnns, fingerprints_vs_gnns_2, MOLTOP, peptides_fingerprints}.

Sets of molecular descriptors, exemplified by Mordred \cite{Mordred}, are also commonly used as molecular fingerprints for molecular property prediction. They include both physicochemical and topological features describing a compound, typically based on its 2D, ``flat'' representation for computational efficiency. However, their feature spaces differ significantly from those of hashed fingerprints, as they extract continuous features and are closer to how chemists typically describe molecules.

\subsection{Graph neural networks}

Most Graph Neural Network (GNN) architectures follow a message-passing framework \cite{GNNs_message_passing}. The initial embedding of each atom consists of elementary chemical descriptors such as the element type or charge. In each GNN layer, an atom receives embeddings from its neighbors (messages) and updates its own embedding accordingly, incorporating this information. Nearly all molecular GNNs also include bond features and embeddings in this process. To obtain a whole-molecule embedding, atom embeddings are aggregated using a readout function, such as a channel-wise average or sum. The message-passing mechanism integrates both functional and structural information about the molecule in an end-to-end learning framework.

\textbf{Graph Isomorphism Network (GIN)} \cite{GIN} is one of the most widely used GNN architectures, as it was shown to be as expressive as the Weisfeiler-Lehman isomorphism test in distinguishing non-isomorphic graphs. GIN uses a two-layer multilayer perceptron (MLP) for atom embedding updates during message passing and employs a sum readout. It is often used by GNN pretraining methods as a backbone.

\textbf{Context Prediction (ContextPred)} \cite{pretraining_GNNs_GIN} was proposed as an SSL pretraining procedure for message-passing GNNs, in particular the GIN backbone. For each atom, a $K$-hop neighborhood subgraph is defined and encoded as a vector by the GNN that is being pretrained. A corresponding context subgraph, located between $r_1$ and $r_2$ hops away, is encoded by a separate GNN. The pretraining objective is binary classification with negative sampling, where the model learns to distinguish correct (positive) neighborhood-context pairs from randomly sampled (negative) ones.

\textbf{GraphMVP} \cite{GraphMVP} combines contrastive and generative SSL, aligning molecular 2D and 3D representations, i.e. topological graphs and conformations. These two views are encoded using GIN and SchNet \cite{SchNet}, respectively. In the contrastive setup, positive pairs consist of a molecule and its conformers, while the generative objective minimizes the variational autoencoder (VAE) loss between the two representations.

\textbf{GraphFP} \cite{GraphFP} introduces graph fragmentation for pretraining, employing both contrastive and predictive SSL. Frequent subgraph mining is used to decompose molecules into fragments. In contrastive learning, fragments and their constituent atoms form positive pairs, while atoms from unrelated fragments serve as negative pairs. The atom-encoding GNN is pretrained to generate atom-level embeddings (which can later be pooled), while a separate GNN encodes entire fragments. A GIN backbone is used for both. In addition, a predictive task is used to classify the presence of fragments, providing a multitask pretraining signal for the atom-encoding network.

\textbf{MolR} \cite{MolR} also uses contrastive learning, but incorporates chemical reaction information. Positive samples are constructed from known reactant-product pairs in reaction databases, while negative samples are created by randomly shuffling these pairs. All molecules are encoded using the same GNN being pretrained, based on the TAG architecture \cite{TAG}.

\textbf{GEM} \cite{GEM} relies on predictive SSL, primarily using 3D molecular conformations. Pretraining tasks include predicting bond lengths, bond angles, and interatomic distances, with parts of the molecule masked as input. Additional auxiliary tasks involve predicting molecular fingerprint bits (MACCS and ECFP) based on the topological graph. Unlike the other architectures described above, which use only topological graphs at inference time, GEM requires 3D conformers as input, and uses a custom GeoGNN backbone to process them. This requirement also makes both pretraining and inference computationally expensive due to the need for conformation generation.

\textbf{MolE} \cite{MolE} uses contrastive SSL, generating molecule views by subgraph masking. A mask seed atom is selected at random, and then its neighbors (and their neighbors, and so on) are masked until $25\%$ of atoms have been masked. Two such augmented masked views coming from the same molecule form a positive pair for contrastive learning, and are encoded with a GIN backbone. Barlow Twins loss \cite{BarlowTwins} is minimized to bring positive samples close together.

\textbf{CheMeleon} \cite{Chemeleon} is based on the D-MPNN architecture \cite{DMPNN}, rather than GIN. It reformulates the message-passing paradigm to exchange embeddings between bonds, rather than atoms, and performs asymmetrical, directed operations, which aim to reduce noise. It is pretrained based on the multitask regression (MTR) paradigm, predicting values of Mordred descriptors \cite{Mordred} for the molecule. Target values are masked during pretraining to avoid overfitting, with only a random subset being predicted for each molecule in a batch.

\subsection{Graph transformers}

Graph transformer architectures extend self-attention mechanisms to molecular graphs, replacing or augmenting the localized message-passing of GNNs with global attention layers. Atoms are treated as tokens, and edges are encoded as pairwise biases injected into the attention scores, allowing information to propagate between any two atoms within a single layer. Compared to message-passing GNNs, this design more efficiently captures long-range dependencies and enables the incorporation of rich edge or distance features through additive or multiplicative attention biases.

\textbf{GROVER} \cite{GROVER} is a hybrid transformer-GNN model in which the attention heads are biased by chemoinformatics-derived edge features. It is pretrained using two self-supervised tasks: contextual property prediction and motif prediction. The former involves classifying the local neighborhood of masked atoms and bonds, and the latter is classification of structural motifs, i.e. multitask classification (MTC) on the level of a whole graph. The graph-level embedding is obtained by concatenating a [CLS] token with pooled atomic embeddings.

\textbf{MAT} \cite{MAT} introduces distance-aware attention based on 3D conformations, incorporating adjacency and shortest-path kernels into the query-key dot product. Pretraining combines masked atom and bond prediction, analogous to masked language modeling (MLM) in NLP, encouraging the encoder to learn the molecular topological context.

\textbf{R-MAT} \cite{RMAT} builds on MAT by introducing relative positional encodings derived from graph distances and ring memberships, which improves robustness to molecule size and graph sparsity. Inspired by GROVER, R-MAT uses subgraph masking as the initial pretraining objective, making it more challenging than the single-atom masking of MAT. The second pretraining phase is based on multitask regression (MTR), in which the model predicts 200 physicochemical molecular properties computed with RDKit.

\textbf{Uni-Mol} \cite{UniMol} and its extended version, \textbf{Uni-Mol2} \cite{UniMol2}, apply graph transformers to 3D conformers using SE(3)-equivariant attention mechanisms. The encoder is jointly pretrained to reconstruct masked atoms, predict pairwise distances and angles, and distinguish positive from negative conformer–SMILES pairs. These objectives produce embeddings that integrate both molecular topology and 3D geometry.

\subsection{Text transformers}

Molecules serialized as SMILES \cite{SMILES} or SELFIES \cite{SELFIES} strings can be processed directly using NLP-inspired architectures. Although pretraining can incorporate chemistry-specific biases, it is not strictly necessary, and the simplicity of NLP-derived pretraining is often a notable advantage of these models. In encoder-decoder architectures, only the encoder is used after pretraining to compute molecular embeddings.

\textbf{Continuous Data-Driven Descriptors (CDDD)} \cite{CDDD} is an RNN-based architecture. It was trained using a purely text-based translation task, from random SMILES to canonical SMILES, calculated with RDKit \cite{RDKit}.

\textbf{Chemformer} \cite{Chemformer} is based on the BART encoder-decoder transformer \cite{BART}, pretrained using a denoising SSL objective. Given a SMILES string, it is first randomized (e.g. by starting at a different atom) and some tokens are randomly masked; the model is then trained to reconstruct the original SMILES.

\textbf{MolBERT} \cite{MolBERT} is an encoder-only transformer based on the original BERT architecture. It is pretrained using two objectives: masked language modeling (MLM) and multitask regression (MTR), the latter predicting 200 2D RDKit descriptors. This dual objective is intended to combine contextual learning from SMILES with molecular-level inductive biases.

\textbf{ChemBERTa} \cite{ChemBERTa} uses a RoBERTa-style encoder-only architecture. The authors explore two distinct pretraining variants: one based solely on MLM and another based solely on MTR. The MTR-only variant, predicting 200 2D RDKit descriptors, is reported to be simpler to train compared to MolBERT, and achieves better performance than both the MLM variant and MolBERT's combined objective.

\textbf{MoLFormer} \cite{MolFormer} is another encoder-only SMILES transformer. Its authors aim to further simplify pretraining by relying entirely on a large pretraining corpus (derived from PubChem and ZINC) and standard NLP procedures, rather than domain-specific architectural changes. The model is based on BERT, with modifications including linear attention and rotary position embeddings, and is pretrained using MLM.

\textbf{SimSon} \cite{SimSon} uses an encoder-only backbone and is pretrained by contrastive SSL. Randomized SMILES strings of the same molecule form positive pairs, while strings from different molecules form negative pairs. Despite using such a simple pretraining procedure and a much smaller dataset, SimSon is shown to be competitive with the aforementioned architectures.

\textbf{ChemFM} \cite{ChemFM} is built on a decoder-only LLaMA \cite{LLaMa} backbone and is pretrained using GPT-style causal language modeling (CLM). It is significantly larger than other models, with 1B- and 3B-parameter variants. The molecule embedding is derived from the output vector of the end-of-sequence ([EOS]) token.

\textbf{ChemGPT} \cite{ChemGPT} adopts a decoder-only architecture based on GPT-Neo \cite{GPTNeo}, and is trained on SELFIES strings using CLM. The authors train multiple model variants up to 1.2B parameters, demonstrating that neural scaling laws hold for chemistry-focused text-based transformers as well.

\textbf{SELFormer} \cite{SELFormer} is based on an encoder-only RoBERTa architecture but uses SELFIES, rather than SMILES. Its authors assume that the more regular syntax of SELFIES facilitates learning and improves the capture of relevant molecular patterns. Pretraining uses the MLM objective.

\subsection{Hybrid models}

\textbf{Mol2vec} \cite{Mol2vec} combines graph-based ECFP descriptors with the text-based Word2Vec approach \cite{Word2Vec} to learn context-aware token representations. It first computes integer ECFP identifiers for circular subgraphs centered on each atom, using radii 0 and 1. These subgraphs are treated as words (tokens) in a sequence. A Word2Vec skip-gram model is then used to pretrain context-aware vector representations for each token. During inference, all subgraph vectors are summed to obtain a whole-molecule embedding.

\textbf{COATI} \cite{COATI} employs a multimodal encoder-decoder architecture, pretrained using CLIP-style \cite{CLIP} multimodal contrastive learning. The 3D molecule conformer is encoded using an E(3)-equivariant GNN \cite{E3GNN}, and its SMILES string with a RoFormer \cite{RoFormer}. The model is pretrained with two objectives: contrastive learning between those two representations (treating embeddings of the same molecule as positive pairs), and autoregressive SMILES reconstruction using a decoder. At inference time, molecular embeddings are obtained from the SMILES encoder.

\textbf{CLAMP} \cite{CLAMP} is a multimodal encoder model trained with supervised contrastive learning based on PubChem bioactivity assay data. In the best-performing variant, each molecule is encoded via the concatenation of two count-based molecular fingerprints (ECFP and RDKit) processed through a two-layer MLP, which serves as the final embedding. Assay descriptions are encoded using CLIP \cite{CLIP}. Positive pairs consist of a bioactive molecule and the description of the assay in which it is active, while negative pairs involve non-bioactive compounds. This design incorporates a bioactivity-specific inductive bias into the fingerprint-based representation.

\section{Methods}

We comprehensively evaluated molecular embeddings using 25 datasets, 7 from MoleculeNet \cite{MoleculeNet} and 18 from TDC \cite{TDC}, as they are among the most widely used benchmarks in molecular property prediction. Our focus is on classification tasks, as classification is the most common approach in virtual screening (binary bioactivity classification) and ADMET prediction. This also allows direct comparison of models using the same metric and easy application of Bayesian statistical tests. The datasets span a broad range of tasks, including ADME, various toxicity endpoints, virtual screening, and more. In particular, TDC datasets include several small ones that reflect real-world ML-assisted drug discovery scenarios, where fine-tuning results in quick and strong overfitting. For a realistic out-of-distribution evaluation, we use a challenging scaffold-based train-test split. We leave expanding our framework to regression, e.g. to MoleculeACE \cite{MoleculeACE} or OpenADMET-ExpansionRx \cite{OpenADMET_ExpansionRx} benchmarks, as well as mixing classification and regression tasks, for future work.

All 28 models described in the previous section were evaluated, including 24 neural networks and 4 molecular fingerprints. For models with multiple variants (e.g. different neural network sizes), all viable variants were implemented, and the best-performing variant (based on BBT posterior $\beta$-ranking) was selected. We extracted frozen embeddings (without fine-tuning) from pretrained neural networks and trained the same classification models on top of these features. This setup, inspired by approaches like DINO in computer vision, ensures a fair comparison by removing confounding factors such as downstream classifier architecture, parameter count, or fine-tuning strategy. Performance is affected foremost by the pretrained model's knowledge and its transferability. For classification, we used Random Forest (RF), Logistic Regression (LR), and k-Nearest Neighbors (kNN). RF and LR are widely used in chemoinformatics because of their strong performance on small datasets. kNN, being instance-based and non-parametric, isolates performance differences to the quality of the embeddings. The hyperparameter tuning was identical across all datasets and models to ensure fairness. See the Supplementary Material for details.

We used the area under the receiver operating characteristic curve (AUROC) as an evaluation metric. AUROC is well-suited for imbalanced datasets, common in molecular ML, and is widely adopted in chemoinformatics, including in MoleculeNet and TDC benchmarks. The ECFP count fingerprint was used as the main baseline, due to its popularity, strong theoretical foundations, and consistently strong practical performance \cite{fp_ecfp}. It also captures more information than its binary variant \cite{scikit-fingerprints}. Atom Pair (AP) and Topological Torsion (TT) fingerprints also used count variants, all implemented with the \texttt{scikit-fingerprints} library \cite{scikit-fingerprints,scikit-fingerprints-2}.

\begin{figure}[H]
    \centering
    \includegraphics[width=0.7\columnwidth]{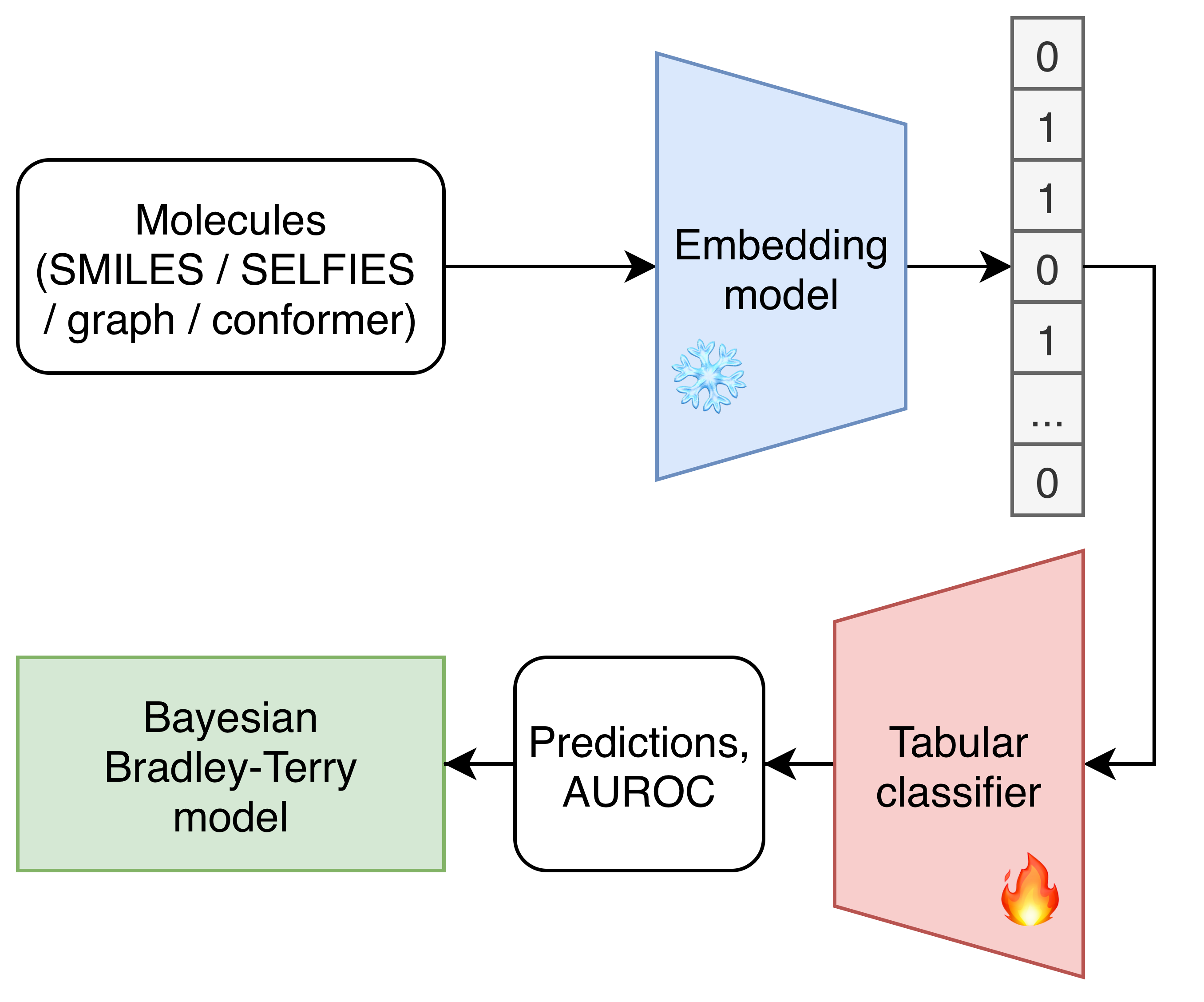}
    \caption{The proposed workflow.}
    \label{workflow_fig}
\end{figure}

Figure \ref{workflow_fig} illustrates our approach. This setup enables the rigorous evaluation of numerous models on a large-scale, diverse benchmark under a fair comparison framework. Results are analyzed with dedicated Bayesian statistical tests: the Bayesian Bradley-Terry model \cite{BBT} to compare all models across all datasets, and the Bayesian hierarchical correlated t-test \cite{bayesian_tests} for the follow-up pairwise comparison of the best model with ECFP on external datasets. Both tests are implemented in the open-source Python library \textit{bayes-testkit} (\url{https://github.com/MLCIL/bayes-testkit}), which we use for all statistical analyses. See the Supplementary Material for details on datasets, model variants, hyperparameter grids, and hardware.

\subsection{CLAMP data leakage}

As noted before, models evaluated in this work were selected based on the availability of open-source code and pretrained weights. Those were used as-is, with the exception of CLAMP, which required retraining due to data leakage present in the available pretrained model. We describe this issue in this section.

CLAMP is pretrained based on supervised contrastive learning, utilizing PubChem bioactivity assays. These labels also appear in multiple evaluation datasets in this work, in many cases directly, e.g. in HIV and MUV datasets from MoleculeNet, and in CYP Veith datasets \cite{CYP_Veith_TDC} from TDC. The original authors removed only test sets from the pretraining data, and only for MoleculeNet, which results in data leakage for our evaluation protocol.

To avoid this problem, and perform fair, leakage-free evaluation, we retrained CLAMP using the original code provided by its authors. The same procedure, hyperparameters, and PubChem 18 assay data were used, and our only change was removing all molecules appearing in our benchmark, based on InChI strings. Throughout the rest of this paper, we use our retrained model, and refer to it simply as ``CLAMP''. Our weights are publicly available on Hugging Face at \url{https://huggingface.co/scikit-fingerprints/benchmarking_models_clamp_refitted}.

\subsection{Bayesian Bradley-Terry (BBT) model}

The benchmarking problem addressed in this work can be framed as a comparison of multiple models on multiple datasets \cite{Demsar}. In such a setting, aggregated metrics alone, such as mean AUROC and mean rank, may not provide a complete picture of model performance \cite{Demsar,BBT}. Mean AUROC can be sensitive to outliers (e.g. datasets unusually easy for certain models), while mean rank does not reflect the magnitude of performance differences. For this reason, Bayesian testing has been recommended for this setting \cite{bayesian_tests}, as it addresses several limitations of frequentist null hypothesis significance tests (NHST). In particular, p-values do not give the probability of a hypothesis, they depend on both the effect size and the sample size, and a non-significant result does not by itself indicate that the models are equivalent \cite{bayesian_tests}. Bayesian tests instead provide the posterior probability of each hypothesis of interest, e.g. that one model is better than another. Further, the region of practical equivalence (ROPE) specifies differences too small to be of practical relevance, which allows a test to return a tie (practical equivalence) as a decision, together with its posterior probability. The BBT model, described below, models $P(i \succ j)$, the posterior probability that model $i$ beats model $j$.

Specifically, we use the hierarchical Bayesian Bradley-Terry (BBT) model \cite{BBT}, which is designed for robust multi-model comparisons on multiple datasets. It provides clear decisions on whether one model outperforms another, the two are practically equivalent, or the evidence is inconclusive, while offering strong theoretical grounding and intuitive probabilistic interpretation. This statistical testing model combines empirical evidence (observed wins), model assumptions (logistic link), and parameter uncertainty into a single probability relevant to the decision. $P(i \succ j)$ is the central quantity reported by this model.

Below, we summarize the BBT model and refer interested readers to the source publications \cite{bayesian_tests,BBT} for detailed descriptions and proofs of statistical properties.

\paragraph{\textbf{Pairwise win counts}}
For every ordered pair of models \((i,j)\), let \(W_{ij}\) be the number of datasets on which model \(i\) outperforms model~\(j\) in a given metric, e.g. AUROC. If the difference in the metric is too small, less than or equal to $0.01$ (i.e. $1\%$), the models are deemed practically equivalent, adding $1$ to both \(W_{ij}\) and \(W_{ji}\) (known as the add rule for ties). We denote the total number of comparisons by $N_{ij} \;=\; W_{ij} + W_{ji}$.

\paragraph{\textbf{Likelihood}}
The win counts are modeled with a binomial Bradley-Terry likelihood:
\begin{equation}
  W_{ij} \;\sim\; \mathrm{Binomial}\!\Bigl(N_{ij},\,\pi_{ij}\Bigr), 
\end{equation}
\begin{equation}
  \pi_{ij} \;=\; 
  \frac{\exp(\beta_i)}{\exp(\beta_i)+\exp(\beta_j)}
  \;=\;
  \bigl(1 + e^{-(\beta_i-\beta_j)}\bigr)^{-1}
  \label{eq:bbt-likelihood}
\end{equation}

\paragraph{\textbf{Priors}}
The latent ability \(\beta_i\) of each model, i.e. its true overall performance, is given a hierarchical Gaussian prior with shared scale~\(\sigma\):
\begin{equation}
  \beta_i \;\sim\; \mathcal{N}\!\bigl(0,\sigma^{2}\bigr),
\end{equation}
\begin{equation}
  \sigma \;\sim\; \mathrm{LogNormal}\!\bigl(0, 0.5^2\bigr),
\end{equation}
\begin{equation}
  \sum_{i=1}^{M}\beta_i \;=\; 0
  \quad\text{(identifiability)}.
  \label{eq:bbt-priors}
\end{equation}

\paragraph{\textbf{Posterior inference}}

To obtain the posterior distribution of the probability that model $i$ is better than model $j$, we sample the joint posterior
\(
p \bigl(\boldsymbol{\beta},\sigma \mid \{W_{ij}\}\bigr)
\)
with Markov chain Monte Carlo (MCMC). Convergence is checked, and posterior predictive checks (PPCs) are used to confirm that the observed win counts are plausible draws from the fitted model.

\paragraph{\textbf{Posterior summaries}}
For each ordered pair \((i,j)\) and each MCMC draw \(s\), we compute

\begin{equation}
  \pi_{ij}^{(s)} \;=\;
\operatorname{logit}^{-1}\!\bigl(\beta_i^{(s)}-\beta_j^{(s)}\bigr) \label{eq:bbt-pi} 
\end{equation}

and report:
\begin{enumerate}
  \item the posterior mean \(\hat{\pi}_{ij}\),
  \item \(P_{\text{in~ROPE}}\), the posterior mass lying within the prespecified
        probability ROPE \([\rho_L,\rho_U]\),
  \item \(P_{\text{above~0.5}}\), the posterior probability that \(\pi_{ij} > 0.5\), i.e. that model \(i\) truly
        outperforms model~\(j\).
\end{enumerate}
Global latent abilities are summarized by the posterior means
\(\hat{\beta}_i=\mathbb{E}[\beta_i]\), which induce a unique ranking while retaining full pairwise uncertainty.

\paragraph{\textbf{Decision rule and model ranking}}
Let \([\rho_L,\rho_U]\) be the probability-space region of
practical equivalence (ROPE). It is defined directly on the win probability \(\pi_{ij}\) and is therefore independent of the scale of the model quality metric (e.g. AUROC). The values of $\rho_L$ and $\rho_U$ control how conservative the test is towards claims of difference: a wider ROPE declares more pairs of models practically equivalent and fewer pairs better or worse.

BBT makes the decision about the relative quality of two models as follows:
\begin{itemize}

    \item if $P_{\text{in~ROPE}} \ge 0.95$, models are equivalent
    \item if $P_{\text{in~ROPE}} < 0.95 \wedge P_{\text{above~0.5}} \ge 0.95$, then model $i$ is better than $j$
    \item if $P_{\text{in~ROPE}} < 0.95 \wedge P_{\text{above~0.5}} \le 0.05$, then model $i$ is worse than $j$
    \item otherwise, evidence is inconclusive, and no decision can be made
\end{itemize}

In addition, we report decisions under the strong interpretation of BBT \cite{BBT}, which are based on the posterior mean win probability $\hat{\pi}_{ij} = \mathbb{E}[\pi_{ij} \mid D]$, using the thresholds proposed in the original work \cite{BBT}:
\begin{itemize}
    \item if $0.45 \le \hat{\pi}_{ij} \le 0.55$, models are equivalent
    \item if $\hat{\pi}_{ij} > 0.70$, then model $i$ is better than $j$
    \item if $\hat{\pi}_{ij} < 0.30$, then model $i$ is worse than $j$
    \item otherwise, evidence is inconclusive, and no decision can be made
\end{itemize}

This rule uses no additional criterion based on posterior spread, and its equivalence region is narrower than the ROPE of the main analysis.

Since each decision concerns a single pair of models, practical equivalence is not transitive. By Equation~\ref{eq:bbt-pi}, $\pi_{ij}$ lies within the ROPE if and only if $\operatorname{logit}(\rho_L) \le \beta_i - \beta_j \le \operatorname{logit}(\rho_U)$, which for the ROPE $[0.35, 0.65]$ used in this work corresponds to $|\beta_i - \beta_j| \le 0.62$. Two models may therefore each be practically equivalent to a third model, while the difference between their own abilities exceeds this bound. In such a case, model $i$ is practically equivalent to model $j$, and model $j$ to model $k$, while model $i$ is better than model $k$.

In addition to the pairwise model comparisons, BBT allows ordering the models by their posterior mean of $\beta_i$. This yields a unique global ranking while retaining pairwise posteriors for uncertainty-aware claims.

\subsection{Step-by-step BBT protocol}

Here, we summarize the procedure of Bayesian statistical testing with BBT for our benchmarking protocol.

We follow the recommendations of \cite{BBT} for model interpretation, but widen the ROPE to $[0.35, 0.65]$: if the posterior of the win probability $\pi_{ij}$ lies within this interval with probability at least $0.95$, we consider the two models practically equivalent. This reduces false claims that one model is better than another when evaluating many models, while making practical equivalence easier to declare.

\begin{enumerate}
    \item \textbf{Compute per-dataset scores}: For each dataset and model, compute AUROC under the shared evaluation protocol (frozen embeddings + fixed classifier head).
    \item \textbf{Pairwise comparisons within the dataset}: For each dataset and each pair of models $(i,j)$, declare a win for the model with the higher AUROC (or a tie if the difference is $\leq1\%$).
    \item \textbf{Build the win/loss table}: Sum wins across datasets for every ordered pair $(i,j)$ and convert ties with the add policy (each tie contributes $1$ to both sides).
    \item \textbf{Fit the BBT}: Use the binomial/logistic likelihood with hierarchical priors $\beta_i \sim \mathcal{N} \bigl(0, \sigma^{2}\bigr)$, $\sigma \sim \mathrm{LogNorm}\!\bigl(0,0.5^2\bigr)$, and sample via MCMC to obtain posterior draws of $\{\beta_i\}$. Check convergence and perform PPC.
    \item \textbf{Summarize pairwise posteriors}: For every pair $(i,j)$, transform draws of $\beta_i - \beta_j$ to draws of $P(i \succ j)$, report the posterior mean, $P_{\text{in~ROPE}}$, and $P_{\text{above~0.5}}$.
    \item \textbf{Make decisions with ROPE}: if $P_{\text{in~ROPE}} \ge 0.95$, declare models $i$ and $j$ equivalent. Otherwise, if $P_{\text{above~0.5}} \ge 0.95$, declare $i$ better, and if $P_{\text{above~0.5}} \le 0.05$, declare $i$ worse. Otherwise, declare the evidence inconclusive.
    \item \textbf{Aggregate and rank}: Rank models by posterior mean $\beta_i$ (their latent ability), and report pairwise probabilities alongside the ranking to convey uncertainty.
\end{enumerate}

\subsection{Selection of model variants}

Many architectures are available in several variants, with different model sizes or pretraining objectives. To compare architectures rather than individual checkpoints, we represent each architecture by a single variant, selected using the BBT posterior. 

First, we apply the protocol above to all 53 variants, using the best prediction head for each dataset, and represent each architecture by its variant with the highest posterior mean $\beta_i$. For molecular fingerprints, this always selects the count variants, which we also fix by design. Second, we refit the BBT model on the 28 selected variants only. All comparisons reported in the ``Results'' section come from this second fit. We do not reuse the posterior of the first fit, because variants of the same architecture produce strongly correlated results. The BBT likelihood would count these as independent comparisons, making the posterior overconfident. 

We note that selecting and evaluating variants on the same datasets may slightly favor architectures with many variants. However, reading the same comparisons from the 53-variant posterior leads to identical decisions for the $35$-$65$\% ROPE, except for Mol2vec, which is borderline: practically equivalent to ECFP in the 53-variant posterior ($P_{\text{in~ROPE}} = 0.953$), and worse in the refit ($0.906$).

\subsection{External datasets validation}
\label{sec:external_datasets}

In addition to commonly used MoleculeNet and TDC datasets, we include 11 further datasets for external validation of the best models. Since most models included in this work were evaluated on those two benchmarks, there is always a possibility of implicit bias, e.g. in their architectural choices. Datasets not used previously provide an additional layer of evaluation of generalization in novel drug design.

We selected 11 datasets from medicinal chemistry and toxicology, each one having at least 500 compounds and offering classification labels: 9 binary and 2 with three classes (DILIrank 2.0 and NICEATM). Selection was based on good descriptions of data sources and preprocessing protocol, using classification by design (rather than discretizing the regression labels ourselves), and overall high quality of the data provided. As many datasets in our core benchmark cover virtual screening scenarios, here we focus on ADMET and toxicity endpoints, as they are typically small and would benefit most from transfer learning offered by pretrained neural networks.

\begin{table}[htbp]
\centering
\begin{tabular}{|l|l|p{0.48\linewidth}|}
\hline
\textbf{Dataset} & \textbf{Reference} & \textbf{Endpoint} \\
\hline
NICEATM Acute Oral Toxicity & \cite{dataset_external_NICEATM, dataset_external_NICEATM_2} & Acute oral rat toxicity: non-toxic (0), intermediate (1), very toxic (2)  \\
\hline
Rat carcinogenicity & \cite{dataset_external_rat_carcinogenicity} & In vivo classification as carcinogen or non-carcinogen \\
\hline
DILIrank 2.0 & \cite{dataset_external_dilirank_2} & Drug-induced liver injury: none (0), low (1), high (2) \\
\hline
B3DB & \cite{dataset_external_B3DB} & Blood-brain barrier penetration, $\log BB$ threshold $-1$ \\
\hline
BSEP & \cite{dataset_external_bsep} & Bile salt export pump inhibition, $\mathrm{pIC}_{50} = 4$ threshold \\
\hline
Mitochondrial toxicity & \cite{dataset_external_mitochondrial_toxicity} & Toxicity to mitochondria \\
\hline
Skin sensitization & \cite{dataset_external_skin_toxicity} & Induction of allergic contact dermatitis \\
\hline
Bioavailability & \cite{dataset_external_bioavailability} & Human oral bioavailability $F \geq 50\%$ \\
\hline
Eye corrosion & \cite{dataset_external_eye_irritation_corrosion} & Eye corrosion per GHS H314/H318 classification \\
\hline
Eye irritation & \cite{dataset_external_eye_irritation_corrosion} & Eye irritation per GHS H319 classification \\
\hline
Human photosensitizers & \cite{dataset_external_photosensitizers} & Induction of photosensitization in humans \\
\hline
\end{tabular}
\caption{Selected external datasets used for evaluation.}
\label{tab:external_datasets}
\end{table}

In each case, we use scaffold 5-fold cross-validation \cite{scaffold_kfold_cv}. Molecules are grouped by their Bemis-Murcko scaffold \cite{bemis_murcko}, and the groups are assigned greedily, largest first, to the currently smallest fold. Since most of those datasets do not provide any standardized train/test split, this approach allows us to apply a uniform evaluation protocol, and a more thorough evaluation of model performance than a single split. Most importantly, it enables the use of more complex statistical methods, namely the Bayesian hierarchical correlated t-test, described in the next section. For the two 3-class datasets, we use AUROC macro-averaged over one-vs-rest comparisons.

See the Supplementary Material for detailed descriptions of those datasets. We also describe preprocessing steps there, such as removing molecules with SMILES parsing errors, duplicated SMILES, or ambiguous labels.

\subsection{Bayesian hierarchical correlated t-test} \label{sec:t_test}

To further evaluate the top-performing models against the ECFP baseline, we use the Bayesian hierarchical correlated t-test \cite{bayesian_tests}. Unlike BBT, which only counts wins on each dataset, this test estimates the size of the expected difference between two models, using the scores from every fold of a $k$-fold cross-validation on each dataset. The test operates on differences between two models, $x_{ij} = score_{left} - score_{right}$ on the $i$-th dataset and $j$-th fold, where positive values mean the left model is better. The vector of differences on the $i$-th dataset is assumed to follow a multivariate normal distribution, $\mathbf{x}_i \sim MVN(\mathbf{1}\mu_i, \Sigma_i)$, where $\mu_i$ is the mean difference between the models on that dataset. The covariance matrix $\Sigma_i$ has diagonal elements $\sigma_i^2$, the variance of the differences, and off-diagonal elements $\rho\sigma_i^2$, where $\rho$ is the correlation between folds caused by their overlapping training sets. By default $\rho = \frac{1}{k}$, following the Nadeau-Bengio heuristic \cite{nadeau-bengio-heuristic}. The mean differences of all datasets are assumed to be drawn from a common Student's $t$ distribution, see Equation \ref{eq:ttest-per-dataset}.

\begin{align}
    \mu_i    &\sim t(\mu_0, \sigma_0, \nu) \nonumber \\
    \sigma_i &\sim \mathrm{unif}(0, \bar{\sigma})
    \label{eq:ttest-per-dataset}
\end{align}

Here, $\bar{\sigma}$ is $1000$ times the average sample standard deviation of the differences within a dataset. Since the per-dataset means $\mu_i$ are drawn from a common distribution, the model estimates them jointly rather than independently. As a consequence, each estimate of $\mu_i$ is shrunk towards the population mean $\mu_0$, and the degree of shrinkage increases with the variance of the differences observed on the given dataset. This shrinkage acts as a form of regularization: it reduces the variance of the per-dataset estimates attributable to a particular data split, and limits the influence of individual datasets with high variance on the estimate of $\mu_0$. In addition, the heavy tails of the Student's $t$ distribution make the model robust to datasets whose differences deviate substantially from the others. For a detailed discussion, we refer the reader to \cite{bayesian_tests}. The remaining parameters are given hyperpriors following \cite{bayesian_tests}, see Equation \ref{eq:ttest-hyperpriors}.

\begin{align}
    \mu_0    &\sim \mathrm{unif}(-1, 1) \nonumber \\
    \sigma_0 &\sim \mathrm{unif}(0, \bar{\sigma}_0) \nonumber \\
    \nu      &\sim \mathrm{Gamma}(\alpha, \beta) \nonumber \\
    \alpha   &\sim \mathrm{unif}(0.5, 5) \nonumber \\
    \beta    &\sim \mathrm{unif}(0.05, 0.15)
    \label{eq:ttest-hyperpriors}
\end{align}

Here, $\bar{\sigma}_0$ is $1000$ times the standard deviation of the per-dataset mean differences. The bounds of $\mu_0$ suit metrics bounded in $[0, 1]$, such as AUROC, while the bounds of $\sigma_i$ and $\sigma_0$ are deliberately loose, so that the data, rather than the priors, determine the variances. The posterior distribution is sampled with MCMC, and convergence is checked in the same way as for BBT.

The test makes its decision about the mean difference $\mu_{new}$ on a new, unseen dataset from the same population. Given a ROPE $[-r, r]$, each posterior draw of $(\mu_0, \sigma_0, \nu)$ yields three probabilities: that the left model is practically better on the new dataset, $\theta_L = P(\mu_{new} > r)$; that the models are practically equivalent, $\theta_E = P(-r \le \mu_{new} \le r)$; and that the right model is practically better, $\theta_R = P(\mu_{new} < -r)$. We report $P_L$, $P_E$ and $P_R$, the fractions of posterior draws in which $\theta_L$, $\theta_E$ or $\theta_R$, respectively, is the largest of the three \cite{bayesian_tests}. These sum to one, and if any one of them is at least $0.95$, we make the corresponding decision; otherwise, the evidence is inconclusive. We use $r = 0.01$, i.e. $1\%$ of AUROC, the same as the tie threshold of BBT. In addition, the posteriors of $\mu_i$ show the difference between the models on each dataset after accounting for the information shared across datasets.

The Nadeau-Bengio heuristic assumes random, exchangeable folds. The external validation datasets used for this comparison are evaluated with scaffold-based $k$-fold cross-validation (see Section ``External datasets validation''), which can give folds of slightly unequal size. As such, the true correlation $\rho$ is unknown, and we repeat the test for $\rho \in \{0, 0.001, 0.01, 0.05, 0.1, 0.2\}$. Smaller values of $\rho$ make the per-dataset estimates more precise, and therefore make it easier to detect a difference between the models. At $\rho = 0$, the folds are treated as independent, which is the least conservative setting. Larger values of $\rho$ assume a stronger correlation between folds, and thus less information within each dataset, which widens the posterior and makes a definite decision harder. A conclusion that holds across this whole range therefore does not depend on the heuristic. The results for all values of $\rho$ are reported in Section ``External CheMeleon validation''.

\section{Results}

Table~\ref{tab:mean_rank} reports the mean rank and mean AUROC of each architecture, represented by its variant with the highest posterior mean $\beta_i$ in the BBT fit on all variants (see the Methods section). Scores and rankings of all model variants are available in the Supplementary Material.

The key finding is that only four pretrained neural networks rank above the ECFP fingerprint in mean rank: CheMeleon, R-MAT, MolBERT and ChemBERTa. The pretrained CheMeleon GNN is the top-scoring model both in mean rank and in mean AUROC. All four architectures share a pretraining strategy that sets them apart from the other neural networks: multitask regression (MTR) of molecular descriptors computed with RDKit or Mordred. The remaining models are pretrained with masking, contrastive, generative or substructure classification objectives. For ChemBERTa, the MTR variant outperformed the MLM variant for every model size and classifier (see the Supplementary Material), and thus it is also used here.

\begin{table}[H]
    \centering
    \begin{tabular}{l|c|c}
        \textbf{Model} & \textbf{Mean rank} $\downarrow$ & \textbf{Mean AUROC} $\uparrow$ \\
        \hline\hline
        CheMeleon      &                            5.24 &                        81.17\% \\
        R-MAT          &                            6.52 &                        80.91\% \\
        MolBERT        &                            8.16 &                        80.51\% \\
        ChemBERTa      &                            8.68 &                        80.01\% \\ \hline
        \textbf{ECFP}  & \textbf{8.80}                   & \textbf{79.95\%} \\ \hline
        Mordred        &                            8.92 &                        80.22\% \\
        CDDD           &                            9.30 &                        80.63\% \\
        MAT            &                            9.66 &                        80.14\% \\
        Atom Pair      &                           10.00 &                        79.46\% \\
        MolE           &                           10.30 &                        79.57\% \\
        MoLFormer      &                           11.30 &                        79.80\% \\
        Mol2vec        &                           12.04 &                        79.56\% \\
        CLAMP          &                           13.56 &                        78.61\% \\
        TT             &                           13.92 &                        78.13\% \\
        ChemFM         &                           14.96 &                        78.34\% \\
        COATI          &                           16.04 &                        78.03\% \\
        Uni-Mol2       &                           16.08 &                        78.15\% \\
        MolR           &                           16.28 &                        77.92\% \\
        GraphMVP       &                           17.16 &                        77.32\% \\
        Chemformer     &                           17.36 &                        77.45\% \\
        GraphFP        &                           18.18 &                        76.68\% \\
        Uni-Mol        &                           19.24 &                        76.91\% \\
        GROVER         &                           19.88 &                        75.68\% \\
        ChemGPT        &                           21.86 &                        73.86\% \\
        GEM            &                           22.12 &                        74.36\% \\
        SimSon         &                           22.76 &                        73.52\% \\
        SELFormer      &                           23.44 &                        73.24\% \\
        ContextPred    &                           24.24 &                        72.09\% \\
    \end{tabular}
    \caption{Average model results, with ECFP marked.}
    \label{tab:mean_rank}
\end{table}

\begin{table}[H]
    \centering
    \begin{tabular}{c|c}
        \textbf{BBT decision} & \textbf{Models} \\
        \hline\hline
        Better                & \begin{tabular}[c]{@{}c@{}} CheMeleon \end{tabular} \\ \hline
        \begin{tabular}[c]{@{}c@{}}Practically\\ equivalent\end{tabular} & \begin{tabular}[c]{@{}c@{}} R-MAT, MolBERT, ChemBERTa, \\ Mordred, CDDD, MAT, \\ Atom Pair, MolE, MoLFormer \end{tabular} \\ \hline
        Worse                 & \begin{tabular}[c]{@{}c@{}} Mol2vec, CLAMP, TT, ..., \\ (all other models) \end{tabular} \\
    \end{tabular}
    \caption{Bayesian Bradley-Terry (BBT) model decisions for all models compared to the ECFP baseline, with ROPE $[0.35, 0.65]$.}
    \label{tab:bbt_decisions}
\end{table}

\subsection{Bayesian testing results}

As discussed in the Methods section, mean AUROC and mean rank are insufficient for a proper comparison of multiple models on multiple datasets. Thus, we employ the BBT model and report its results in Table~\ref{tab:bbt_decisions}, with ECFP as the baseline. Following the procedure described in \cite{BBT}, we categorize models as better than ECFP, practically equivalent (within the ROPE), worse, or inconclusive (insufficient evidence). All posterior probabilities are provided in the Supplementary Material. On this benchmark, only CheMeleon is better than ECFP, and no comparison is inconclusive.

Several transformer-based models, including R-MAT, MolBERT and ChemBERTa, are practically equivalent to ECFP, but at a much higher computational cost. A key practical advantage of Bayesian testing is its ability to declare such practical equivalence, which a non-significant result of a null hypothesis test does not imply, while the hierarchical model accounts for multiple comparisons \cite{bayesian_tests}.

Other practically equivalent models include Mordred, CDDD, MAT, Atom Pair, MolE and MoLFormer. All other models are worse than ECFP. This does not mean that these models always perform poorly, but that they are more likely to lose to ECFP than to win against it, with the posterior mean probability of beating ECFP ranging from 0.38 (Mol2vec) to 0.07 (ContextPred). Such insights help guide model selection, reducing engineering effort and computational cost.

We also note that we used a wide ROPE of $35$-$65$\%, which makes practical equivalence easier to declare and shifts borderline comparisons towards a tie. Even so, CheMeleon remains better than ECFP, and the posterior probability supporting each decision against ECFP is at least $99.4\%$. Detailed tables and experiments with other ROPE intervals are available in the Supplementary Material.

As a complementary check, the Supplementary Material reports decisions under the strong interpretation of BBT \cite{BBT}, which are based on the posterior mean win probability. Under it, no model is classified as better than ECFP, and CheMeleon is inconclusive. This warrants a more cautious interpretation of its advantage, consistent with the inconclusive external validation below.

\begin{figure}[H]
    \centering
    \includegraphics[width=\columnwidth]{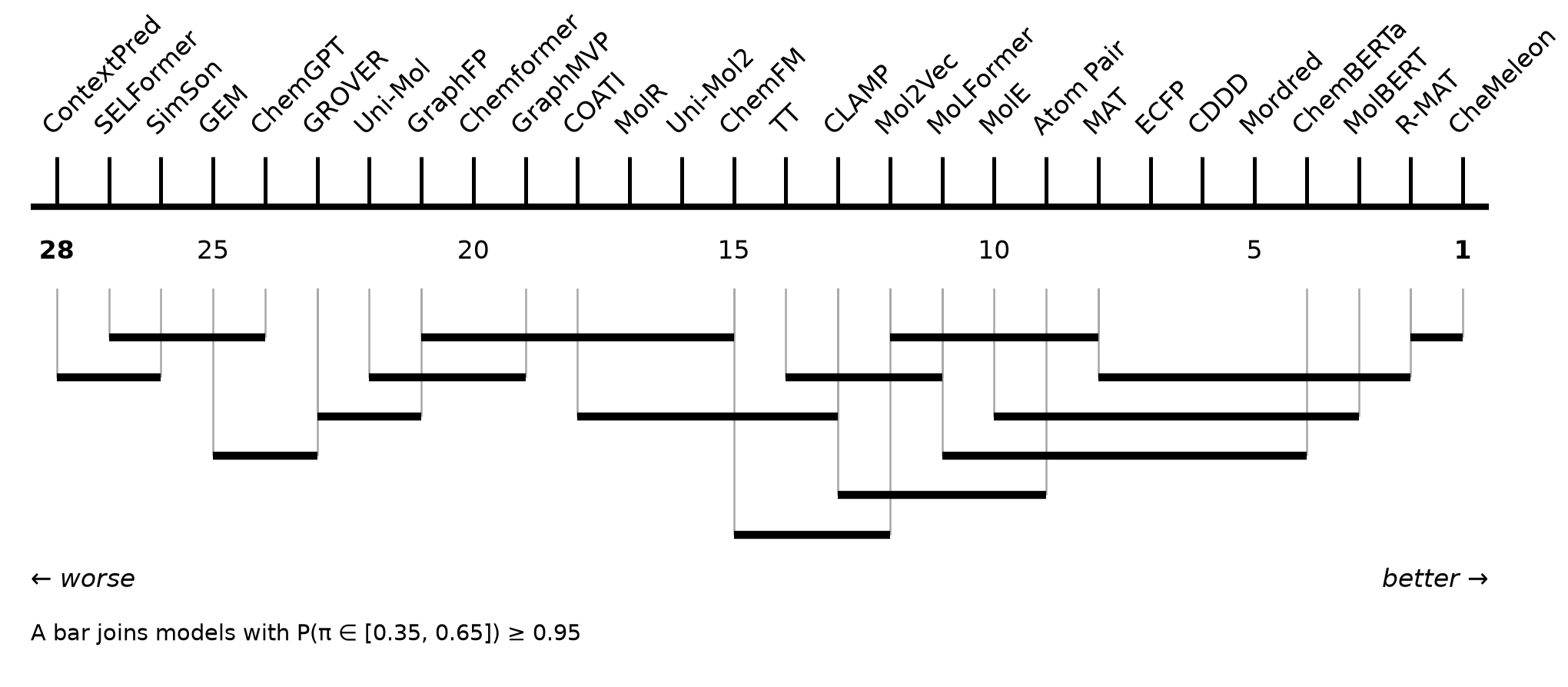}
    \caption{Critical difference diagram, models ordered by BBT posterior mean $\beta_i$. Horizontal bars join groups of pairwise practically equivalent models.}
\label{fig:bbt_cdd}
\end{figure}

Figure~\ref{fig:bbt_cdd} shows the groups of practically equivalent models in a critical difference diagram, adapted from \cite{Demsar}. Models are ordered by their posterior mean $\beta_i$, and a horizontal bar joins each largest group of models that are pairwise practically equivalent, i.e. $P_{\text{in~ROPE}} \ge 0.95$ for every pair in the group. As noted in the Methods section, practical equivalence is not transitive: CheMeleon is practically equivalent to R-MAT, and R-MAT to ECFP, but CheMeleon is better than ECFP.

\subsection{Inter-model win rates}

\begin{figure}[H]
    \centering
    \includegraphics[width=0.7\columnwidth]{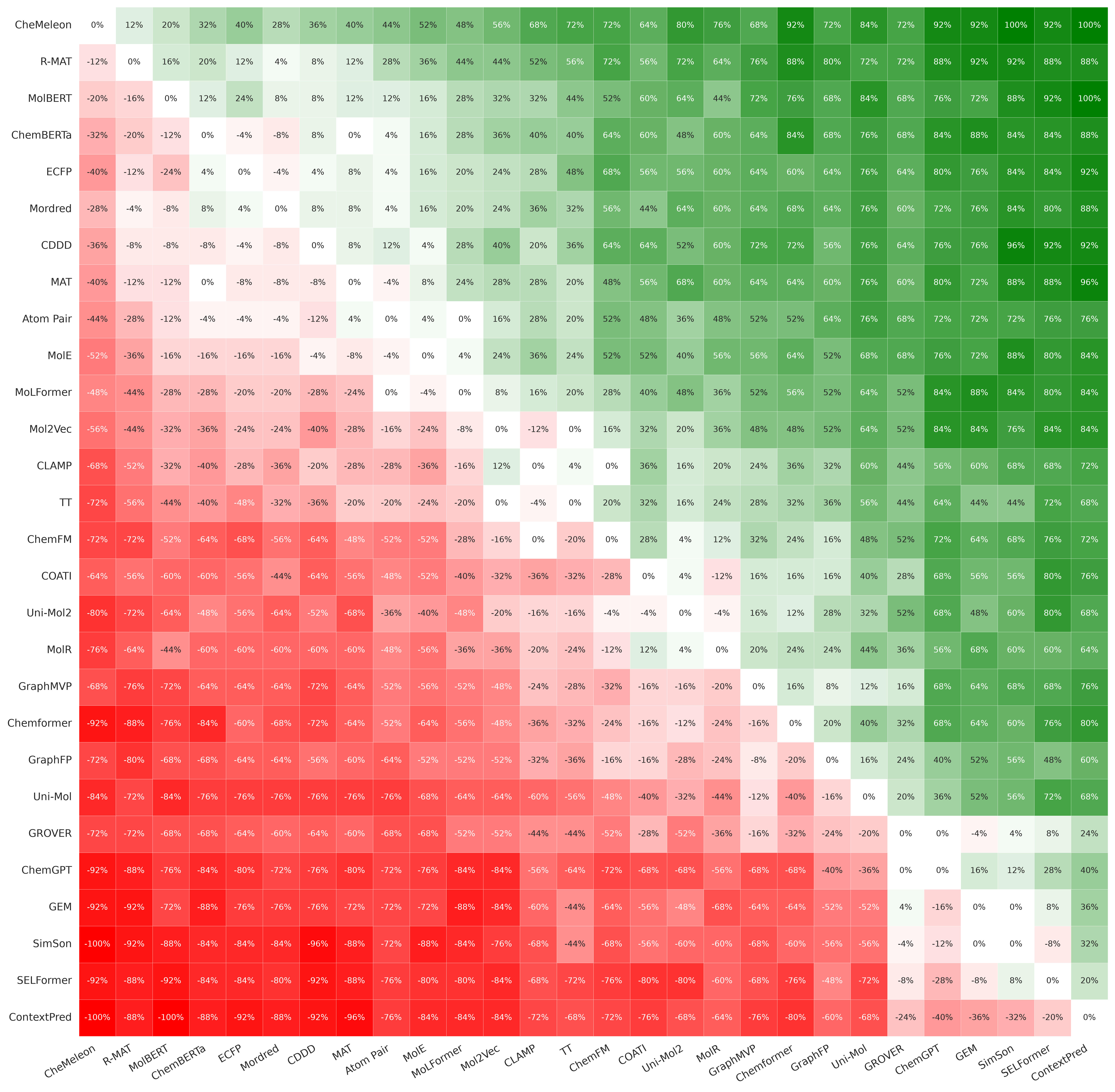}
\caption{Pairwise win ratio between all evaluated models. Green marks wins of the model on the Y axis against the model on the X axis, red means losses (i.e. the model from X axis won more).}
\label{fig:cross_win_rate}
\end{figure}

Figure~\ref{fig:cross_win_rate} provides a detailed analysis of win rates between models. Each cell shows the net win rate of the model in the row against the model in the column, $\frac{1}{n}(n_{\text{wins}} - n_{\text{losses}})$, where $n = 25$ is the number of datasets. Models are ordered by mean rank, as in Table~\ref{tab:mean_rank}. As in BBT, following \cite{bayesian_tests}, we treat AUROC differences below or equal to 0.01 as ties (practical equivalence), and we use the same threshold in all subsequent analyses.

The results align with the Bayesian analysis. Notably, CheMeleon has a positive net win rate against every other model. The smallest margin is against R-MAT, where CheMeleon wins on 3 more datasets than it loses. In contrast, SELFormer, SimSon, GEM, ChemGPT and ContextPred are among the worst-performing models, GEM despite its strong geometric and chemistry-based inductive biases. Both SELFIES-based models, SELFormer and ChemGPT, are among them.

\subsection{Per-dataset wins against ECFP}

\begin{figure}[H]
    \centering
    \includegraphics[width=0.85\textwidth]{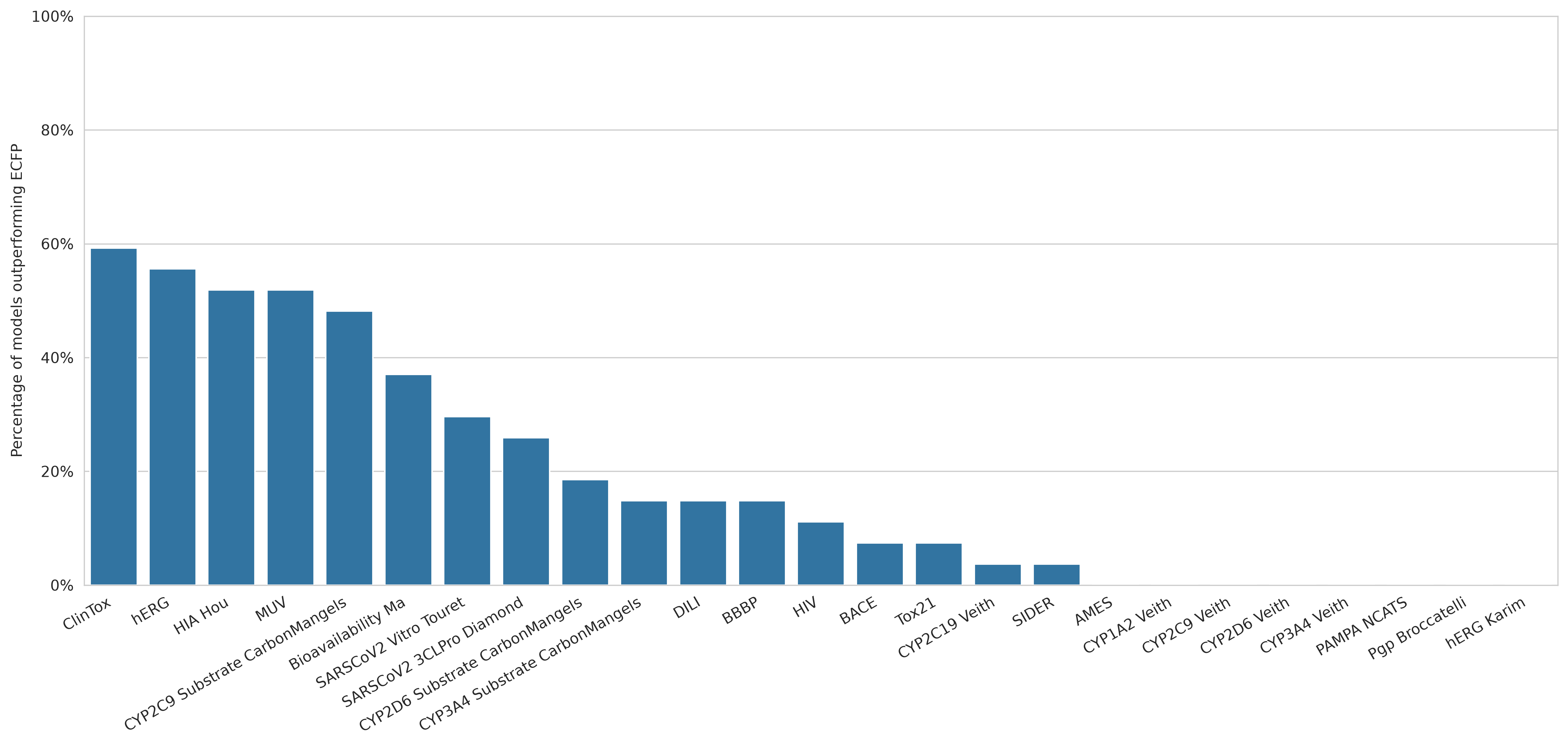}
    \caption{Percentage of models outperforming ECFP (AUROC higher by more than 0.01) on each dataset.}
    \label{fig:dataset_winrate}
\end{figure}

Figure~\ref{fig:dataset_winrate} shows, for each dataset, the percentage of models that outperform the ECFP baseline, i.e. exceed its AUROC by more than 0.01. Notably, on 8 of the 25 datasets no model outperforms ECFP.

We also counted, for each model, the number of datasets on which it won, i.e. achieved the highest AUROC among all models, or had a near-win, i.e. was within 0.01 AUROC of the winner. Table~\ref{tab:close_to_winning} lists the five models with the most wins or near-wins (see the Supplementary Material for all models). Again, only CheMeleon stands out, winning or nearly winning on 12 datasets, almost half of the benchmark, compared to 8 for ECFP, MolBERT, Mordred and R-MAT.

\begin{table}[H]
    \centering
    \begin{tabular}{l|r}
        Model     & Number of datasets \\
        \hline\hline
        CheMeleon &                 12 \\
        ECFP      &                  8 \\
        MolBERT   &                  8 \\
        Mordred   &                  8 \\
        R-MAT     &                  8 \\
    \end{tabular}
    \caption{Number of datasets on which a model achieved the highest AUROC or was within 0.01 AUROC of it, for the five models with the most such datasets.}
    \label{tab:close_to_winning}
\end{table}

\subsection{External CheMeleon validation}

We follow the main study with a deeper analysis of CheMeleon, the only model that BBT found better than ECFP, on the 11 external datasets. While BBT focuses on counting how often one model wins against another, the hierarchical t-test instead estimates the size of the expected AUROC difference between them. Table~\ref{tab:posthoc_raw} shows the AUROC of CheMeleon and ECFP on each external dataset, obtained via 5-fold scaffold cross-validation. Notably, the difference between the models is smaller than their mean standard deviations over folds.

\begin{table}[htbp]
\centering
\begin{tabular}{|l|c|c|}
\hline
\textbf{Dataset} & \textbf{CheMeleon AUROC [\%]} & \textbf{ECFP AUROC [\%]} \\
\hline
NICEATM Acute Oral Toxicity & 80.69 $\pm$ 2.48 & 80.09 $\pm$ 2.37 \\
\hline
Rat carcinogenicity & 68.43 $\pm$ 3.83 & 67.35 $\pm$ 4.49 \\
\hline
DILIrank 2.0 & 70.62 $\pm$ 3.11 & 70.83 $\pm$ 3.38 \\
\hline
B3DB & 91.41 $\pm$ 2.46 & 92.26 $\pm$ 2.38 \\
\hline
BSEP & 86.30 $\pm$ 2.93 & 84.92 $\pm$ 4.85 \\
\hline
Mitochondrial toxicity & 87.00 $\pm$ 2.08 & 83.78 $\pm$ 4.53 \\
\hline
Skin sensitization & 83.25 $\pm$ 2.99 & 81.89 $\pm$ 3.09 \\
\hline
Bioavailability & 75.29 $\pm$ 2.32 & 74.20 $\pm$ 2.32 \\
\hline
Eye corrosion & 94.71 $\pm$ 8.16 & 91.77 $\pm$ 10.48 \\
\hline
Eye irritation & 93.16 $\pm$ 8.89 & 91.02 $\pm$ 13.24 \\
\hline
Human photosensitizers & 72.52 $\pm$ 4.47 & 72.73 $\pm$ 1.48 \\
\hline\hline
\textbf{Mean} & \textbf{82.13} & \textbf{80.98} \\
\hline
\end{tabular}
\caption{Results of the post-hoc comparison on external validation datasets. Mean $\pm$ standard deviation is reported across 5-fold scaffold cross-validation. For each dataset and model, the classification head with the best mean AUROC is used.}
\label{tab:posthoc_raw}
\end{table}

\begin{figure}[H]
    \centering
    \includegraphics[width=0.85\textwidth]{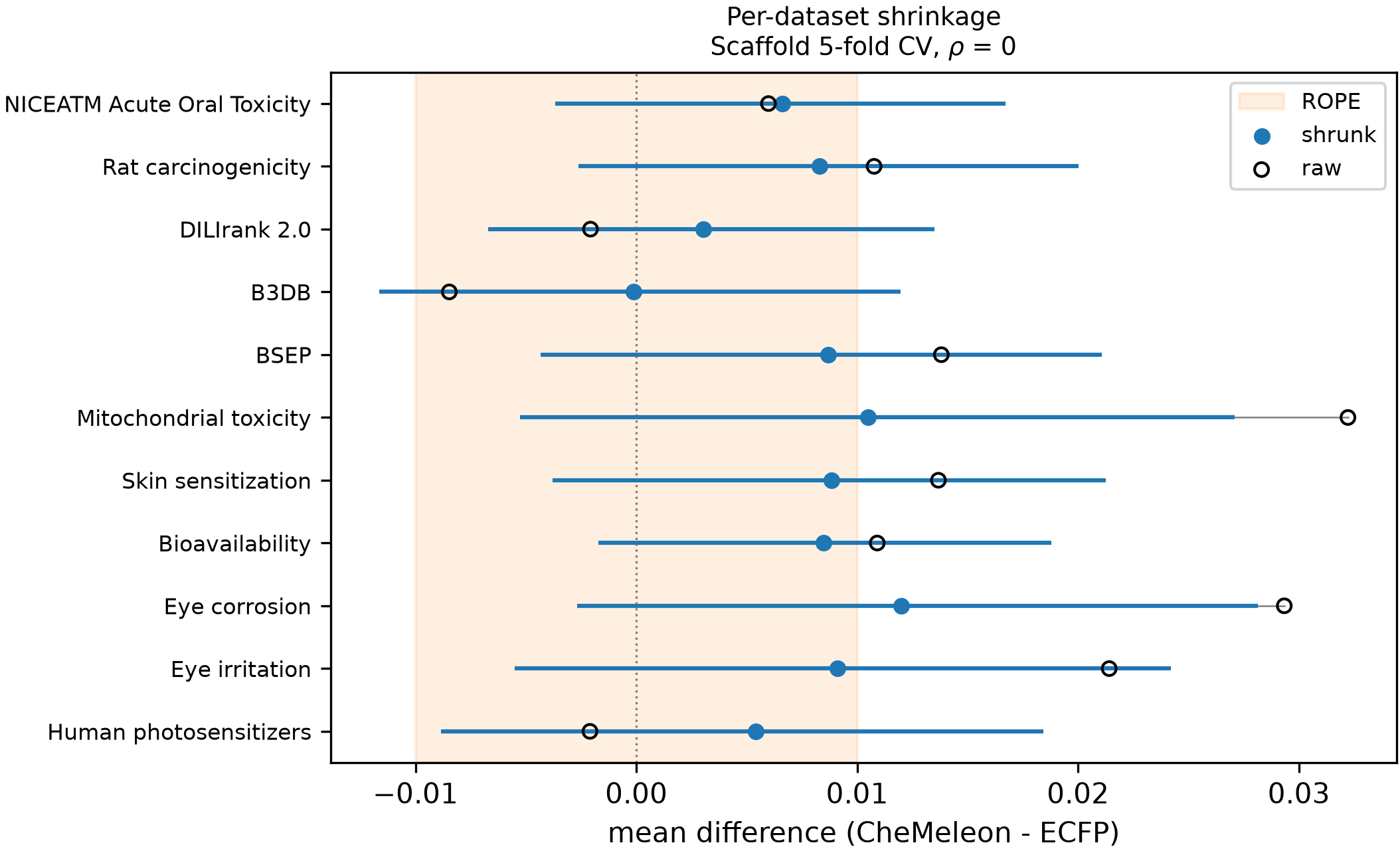}
    \caption{Per-dataset posterior of the mean AUROC difference (CheMeleon $-$ ECFP) from the hierarchical t-test, $\rho = 0$. Filled dots and bars show the posterior mean and 89\% HDI after shrinkage, empty dots the raw mean difference over folds. The shaded area is the ROPE $[-0.01, 0.01]$.}
    \label{fig:posthoc_forest}
\end{figure}

Figure~\ref{fig:posthoc_forest} shows the posterior of the mean difference $\mu_i$ on each dataset. Shrinkage, described in Section ``Bayesian hierarchical correlated t-test'', pulls the per-dataset estimates towards the population mean $\mu_0$, most strongly for datasets with a high variance of differences across folds. Since the fold correlation $\rho$ is unknown for scaffold splits (Section ``Bayesian hierarchical correlated t-test’’), we present the results for $\rho = 0$, i.e. independent folds. This is the least conservative setting: the per-dataset posteriors are the narrowest and the shrinkage is the weakest, which favors detecting differences between the models. Even so, all 89\% highest density intervals (HDIs) include zero. The largest differences in favor of CheMeleon are on eye corrosion, where the raw difference of $+2.93\%$ is shrunk to $+1.20\%$, and on mitochondrial toxicity, shrunk from $+3.22$ to $+1.05$. These are the only datasets with posterior means slightly outside the ROPE. For $\rho = 0.2$, the shrinkage is stronger, e.g. the difference on mitochondrial toxicity is shrunk to $+0.84$, and all per-dataset means lie within the ROPE.

Next, we analyze the posterior of the population mean $\mu_0$, repeating the test for $\rho$ ranging from $0$ to $0.2$, the Nadeau-Bengio default. Table~\ref{tab:posthoc_rho} shows the expected difference between CheMeleon and ECFP $\mathbb{E}[\mu_0]$, its 89\% HDI following \cite{BBT}, and the probabilities that ECFP is practically better, that the models are practically equivalent, and that CheMeleon is practically better.

For all values of $\rho$, the evidence is inconclusive. The most likely outcome is practical equivalence, with probability from $66.2\%$ to $68.1\%$, followed by CheMeleon being practically better, with probability from $31.1\%$ to $33.3\%$. The expected difference $\mathbb{E}[\mu_0]$ ranges from $+0.65\%$ to $+0.74\%$, below the ROPE threshold, and every HDI includes zero.

\begin{table}
    \centering
    \begin{tabular}{r|c|c|c|c|c|l}
        $\rho$ & $\mathbb{E}[\mu_0]$ & 89\% HDI           & P(ECFP) & P(equivalent) & P(CheMeleon) & Decision \\
        \hline\hline
             0 &             +0.0073 & [-0.0004, +0.0155] &   0.005 &         0.662 &        0.333 & Inconclusive \\
         0.001 &             +0.0074 & [-0.0008, +0.0151] &   0.005 &         0.662 &        0.333 & Inconclusive \\
          0.01 &             +0.0073 & [-0.0007, +0.0154] &   0.005 &         0.666 &        0.329 & Inconclusive \\
          0.05 &             +0.0071 & [-0.0012, +0.0156] &   0.006 &         0.681 &        0.313 & Inconclusive \\
           0.1 &             +0.0069 & [-0.0023, +0.0156] &   0.007 &         0.679 &        0.314 & Inconclusive \\
           0.2 &             +0.0065 & [-0.0033, +0.0162] &   0.013 &         0.676 &        0.311 & Inconclusive \\
    \end{tabular}
    \caption{Results of the Bayesian hierarchical correlated t-test, CheMeleon against ECFP, for different values of the fold correlation $\rho$.}
    \label{tab:posthoc_rho}
\end{table}

\begin{figure}[H]
    \centering
    \begin{subfigure}[t]{0.49\linewidth}
        \centering
        \includegraphics[width=\linewidth]{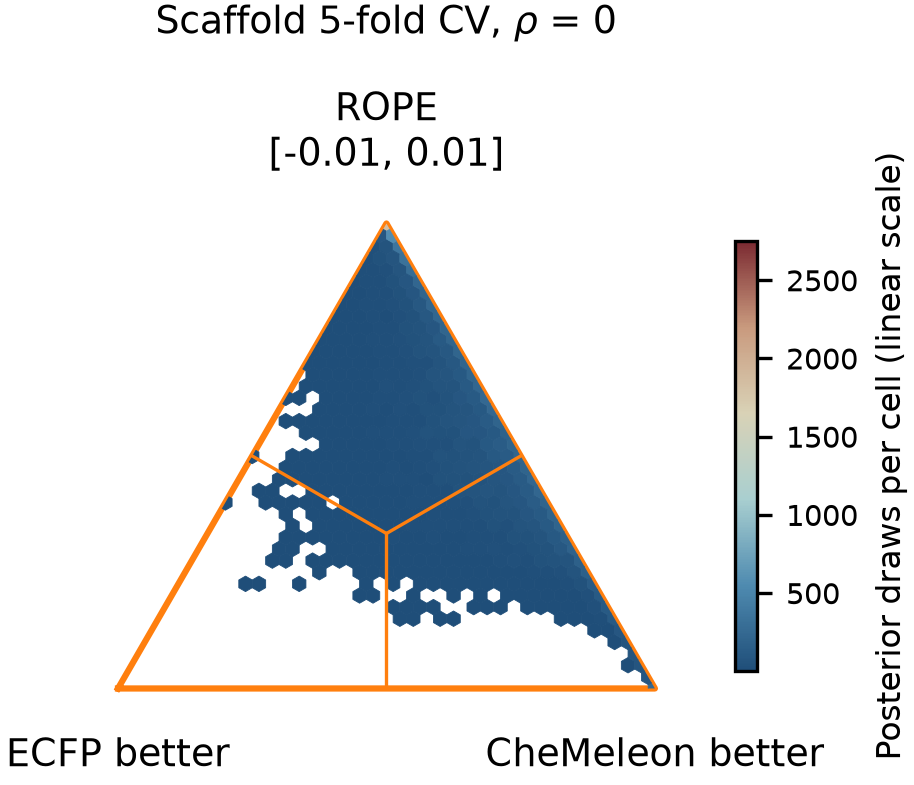}
        \caption{Simplex for $\rho=0$}
        \label{fig:posthoc_simplex_rho0}
    \end{subfigure}\hfill
    \begin{subfigure}[t]{0.49\linewidth}
        \centering
        \includegraphics[width=\linewidth]{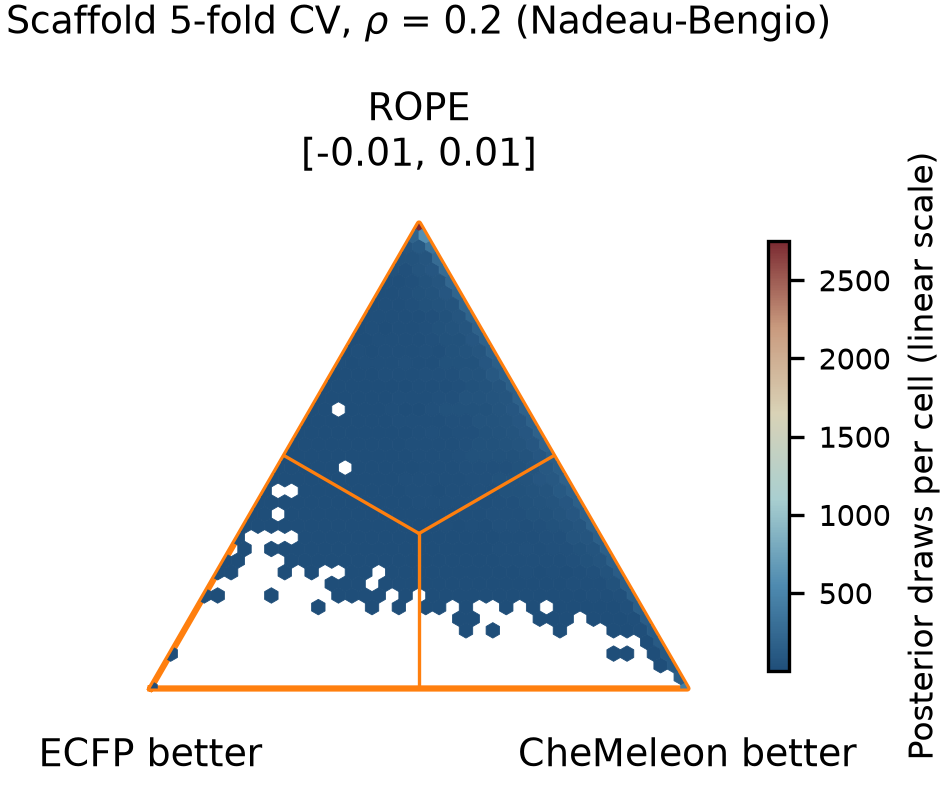}
        \caption{Simplex for $\rho=0.2$}
        \label{fig:posthoc_simplex_rho02}
    \end{subfigure}
    \caption{Posterior predictive distribution of the CheMeleon $-$ ECFP AUROC difference on a new
    dataset, from the hierarchical correlated $t$-test on the scaffold-split folds
    (ROPE $= [-0.01, 0.01]$).}
    \label{fig:posthoc_simplex}
\end{figure}

The simplex plots in Figure~\ref{fig:posthoc_simplex} show the same inconclusiveness. For both $\rho = 0$ and $\rho = 0.2$, posterior draws fall into all three regions of the simplex, concentrated in the equivalence and CheMeleon regions. On a new dataset, CheMeleon will most likely be at least as good as ECFP (probability of at least $98.7\%$ for every $\rho$), but we cannot decide whether it is practically better or equivalent.

Although both tests use the same threshold of 0.01 AUROC, this does not contradict the BBT result, because they apply it differently. BBT applies it to a single score difference on each dataset and counts the outcomes: in head-to-head comparisons, CheMeleon wins against ECFP on 12 of the 25 benchmark datasets, ties on 11 and loses on 2. The same rule applied to the external datasets gives 7 wins, 4 ties and no losses (Table~\ref{tab:posthoc_raw}). The t-test instead applies the threshold to the expected difference on a new dataset, weighting each dataset by how precisely its difference is measured across folds. On all external datasets except eye corrosion, the standard deviation of the per-fold differences exceeds their mean, and on the four datasets where the differences vary least (B3DB, DILIrank 2.0, bioavailability and NICEATM), the mean difference is only $+0.16\%$. Thus, CheMeleon wins against ECFP more often than it loses, as BBT shows, but its expected advantage on a new dataset is most likely below $1\%$ of AUROC.

As a practical takeaway, we conclude that CheMeleon is a strong competitor to the classic ECFP count fingerprint. While it is not clearly superior, it can be expected to give strong results, at least practically equivalent to ECFP. As such, for new QSAR/QSPR projects, we recommend evaluating both count ECFP fingerprints and CheMeleon embeddings.

\subsection{Computational cost}

\begin{figure}[H]
    \centering
    \includegraphics[width=0.85\textwidth]{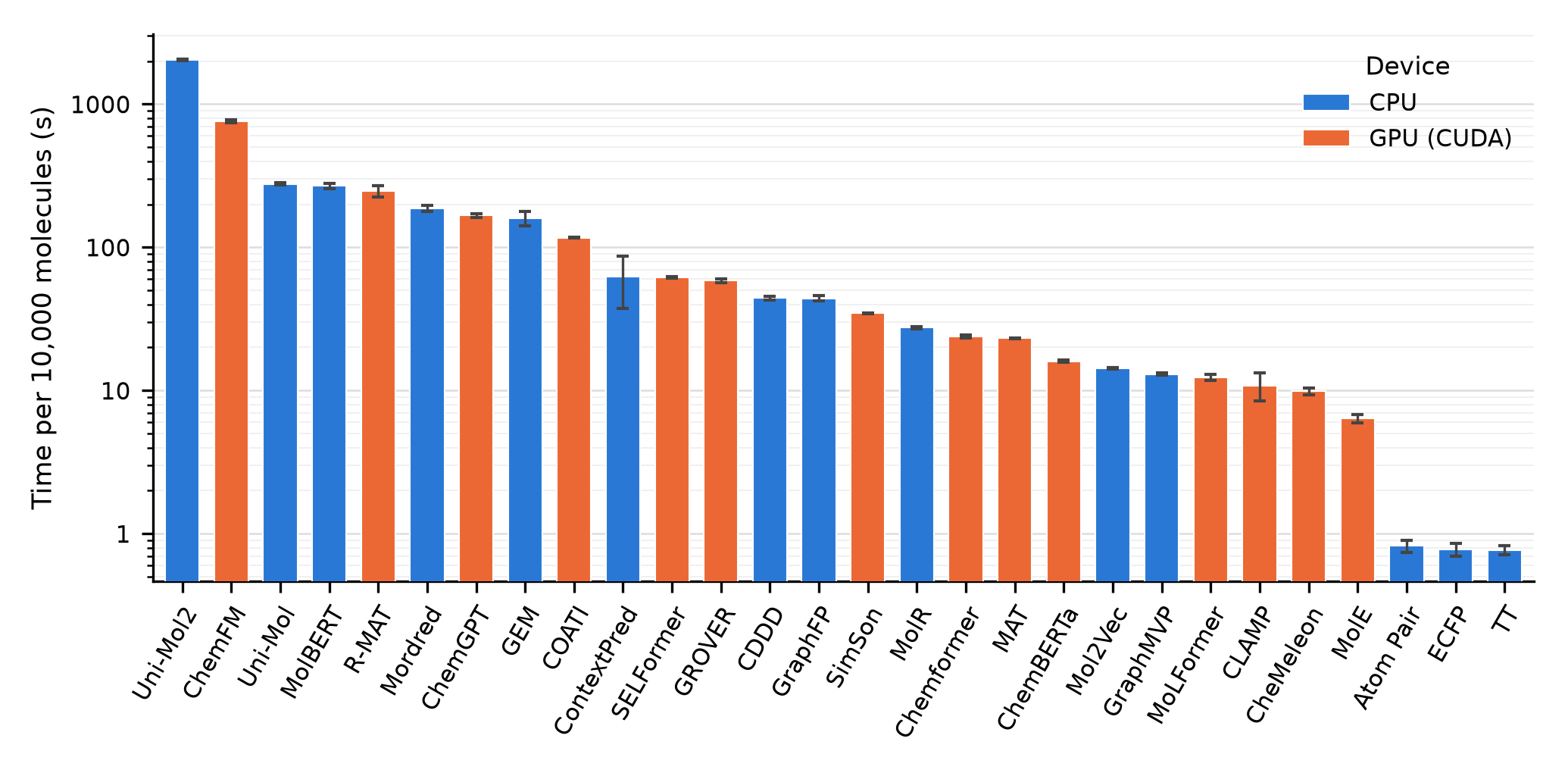}
    \caption{Embedding time of each evaluated model, using its representative variant: mean time to embed 10,000 molecules from MUV, with error bars indicating the standard deviation over batches. Orange marks models run on GPU.}
    \label{fig:compute_times}
\end{figure}

In addition to the quality of the pretrained representations, we also analyze their computational cost. Figure~\ref{fig:compute_times} shows the mean time and standard deviation of embedding 10,000 molecules from the MUV dataset. Models run on GPU, following the setup recommended by their authors, are shown in orange. CPU-based models used 96 threads (75\% of the machine), and GPU-based models a single GPU. Detailed hardware specifications are provided in the Supplementary Material.

All pretrained neural networks are much slower than the topological molecular fingerprints (ECFP, AP and TT), even with GPU acceleration. The fastest neural network, MolE, takes 6.37 seconds per 10,000 molecules, while each of these fingerprints takes less than one second. Table~\ref{tab:runtime_top} shows the computational cost of the four best pretrained models and ECFP. Interestingly, CheMeleon is also one of the fastest pretrained models, taking $9.88 \pm 0.51$ seconds per 10,000 molecules, with only MolE being faster. CheMeleon is also the fastest of the top-performing neural models, about 25 times faster than R-MAT and MolBERT. The full table of computation times is available in the Supplementary Material.

\begin{table}[t]
    \centering
    \begin{tabular}{c|c|c|c}
        \textbf{Model} & \textbf{Device} & \textbf{Mean time [s]} $\downarrow$ & \textbf{Std time [s]} \\
        \hline\hline
        CheMeleon & GPU & 9.88 & 0.51 \\
        R-MAT & GPU & 246.92 & 22.24 \\
        MolBERT & CPU & 268.15 & 11.60 \\
        ChemBERTa & GPU & 15.96 & 0.29 \\
        ECFP & CPU & 0.77 & 0.08 \\
    \end{tabular}
    \caption{Embedding time of the 5 best models by mean rank (Table~\ref{tab:mean_rank}), using each model's representative variant. MUV is embedded in batches of 10{,}000 molecules.}
    \label{tab:runtime_top}
\end{table}

\subsection{Discussion and practical remarks}

Overall, our experiments consistently show that pretrained molecular embedding models based on various neural networks typically fail to outperform the ECFP baseline. This suggests that progress in molecular representation learning, when properly evaluated, is less substantial than commonly claimed.

In fact, only the CheMeleon model performs consistently well. It relies on two chemistry-specific inductive biases: a D-MPNN backbone and MTR pretraining on Mordred descriptors. D-MPNN centers the message-passing procedure on bonds, rather than atoms like typical MPNNs, and performs asymmetric, directed information exchange, which reduces noisy loops \cite{DMPNN}. This is particularly beneficial for chains connected to ring systems, which in regular MPNN designs can suffer from tottering, i.e. messages being sent back and forth through bridging bonds. Mordred descriptors involve both small-scale and large-scale topological information from the molecule, as well as many physicochemical properties. This requires the model not only to understand the structure, but also how it translates to more complex chemical properties of a compound, required for QSAR/QSPR tasks.

Other well-performing models are primarily transformers that also incorporate strong chemistry-specific biases. Those include molecular attention in R-MAT and MAT, and property prediction tasks in MolBERT and ChemBERTa's MTR pretraining.

We contrast this with generic objectives like masked reconstruction (either SMILES- or graph-based) or contrastive learning between purely structural representations, which are used by most models that perform significantly worse than ECFP. Those include e.g. SELFormer, SimSon, GROVER, ChemFM, and many other transformers.

This highlights an important lesson for future development of molecular representation learning: relying solely on large pretraining datasets and generic graph- or NLP-based architectures is insufficient for chemical data. Domain-specific knowledge and tailored model modifications are essential.

Qualitatively, in our experience, reproducibility and code quality strongly correlated with model performance. The best models featured easy-to-run, well-documented code and dependencies. Molecular fingerprint baselines implemented with \texttt{scikit-fingerprints} \cite{scikit-fingerprints,scikit-fingerprints-2} were also very user-friendly. In contrast, models requiring substantial engineering effort to run, such as GraphFP, GEM, or GraphMVP, often delivered subpar results.

Finally, we offer practical recommendations for practitioners in molecular representation learning and chemoinformatics:
\begin{enumerate}
    \item Always include the ECFP count fingerprint as a baseline, paired with a tree-based classifier. Ideally, also test other fingerprints like Atom Pair and Mordred.
    \item Always evaluate CheMeleon, especially when using embeddings without fine-tuning neural networks, e.g. with very small datasets.
    \item Other notable models to try are R-MAT, MolBERT and ChemBERTa. These are relatively easy to set up, but may show variable performance, not necessarily superior to ECFP.
    \item Avoid models that tend to perform poorly and require significant effort to run, such as GraphFP, GEM, and GraphMVP (see Table \ref{tab:mean_rank} and Figure \ref{fig:cross_win_rate}). Use them only if substantial engineering and computational resources are available.
    \item We echo existing literature \cite{bayesian_tests,BBT} recommending Bayesian statistical tests, particularly the Bayesian Bradley-Terry (BBT) model and Bayesian hierarchical correlated t-test, for multi-model, multi-dataset comparisons, as they provide clear aggregation and intuitive interpretation of results.
\end{enumerate}

\section{Limitations and future work}

This work has a few limitations that can be addressed in future work. We consider them minor and unlikely to change our primary conclusions, but addressing them would make future results in this area more robust.

We focused on classification tasks, targeting the AUROC metric. This is a common setting in ADMET and bioactivity classification tasks, particularly under strong measurement noise, where regression can be unreliable. In the future, the presented approach can be extended to regression and also virtual screening, where ranking metrics like BEDROC \cite{BEDROC} are used. In principle, statistical methods from this work should be directly applicable in such mixed benchmarks, but would require setting appropriate ROPE values.

BBT has four specific limitations in our setting. First, it treats the outcomes of all model pairs on a dataset as independent, although they share the same data: a dataset that suits one model can make it win against many others at once, and these correlated wins are counted as separate evidence. Second, BBT has been reported to produce credible intervals that are too narrow and slightly too high probabilities of one model being better than another \cite{BBT}, so our decisions at the $0.95$ threshold may be optimistic. We therefore also report the strong interpretation of BBT, based only on the posterior mean (see Supplementary Material). Under it, CheMeleon still has the highest probability of outperforming ECFP ($0.64$), which lies between the thresholds of practical equivalence and superiority, i.e. CheMeleon is at least as good as ECFP, but not definitely better. Third, the decisions depend on modeling choices: our ROPE is wider than in the original BBT work, which makes equivalence easier to declare, and each tie is counted as a win for both models. However, the original BBT publication reports the tie policies to give similar results \cite{BBT}. Fourth, for reproducibility and comparability with prior work, we use the publicly available scaffold splits of the MoleculeNet and TDC benchmarks, which provide a single train-test split per dataset. Each model therefore obtains a single AUROC score on each dataset, with no estimate of its variance, and a difference above the $1\%$ tie threshold may still result from the particular split, especially on small test sets. Since BBT only counts wins, it cannot account for this uncertainty. The follow-up comparison of CheMeleon and ECFP on external datasets addresses this with scaffold 5-fold cross-validation and the hierarchical t-test, which uses the variance of differences across folds.

The hierarchical t-test assumes that all folds of a dataset have the same variance and the same correlation with each other. Scaffold folds differ in size and class balance, which can lead to different variances, and the Nadeau-Bengio correlation was derived for random folds. Moreover, with 11 external datasets, the population-level parameters of the test are only weakly informed by the data; in particular, the posterior of the degrees of freedom $\nu$ closely follows its prior. We therefore repeated the test for a range of fold correlations and with an alternative prior on $\nu$ \cite{juarez-steel}, and the decision was unchanged in all cases (see Supplementary Material).

Newer embedding methods are constantly being developed, and static benchmarks can quickly become outdated. Developing a live leaderboard for molecular embeddings, akin to MTEB \cite{MTEB} used in NLP, could be beneficial. Alternative tabular approaches are gaining traction in chemoinformatics, particularly pretrained tabular transformers \cite{TabPFN}, and could eventually replace logistic regression and kNN in such an environment.

\section{Conclusions}

We presented a comprehensive benchmark of pretrained molecular representation models. Analysis of 28 models on 25 datasets, based on a fair comparison setting and supported with a Bayesian Bradley-Terry model, challenges perceived progress in this domain and indicates inadequate evaluation rigor in previous works. All models, except CheMeleon, are either worse than or practically equivalent to the baseline ECFP fingerprint on molecular property prediction tasks. Furthermore, the representations learned by most GNNs without strong domain-specific pretraining were generally of significantly worse quality than those of other approaches. Additional validation on external datasets, using the Bayesian hierarchical correlated t-test on 5-fold scaffold cross-validation, shows that CheMeleon embeddings are highly competitive with ECFP, although not clearly superior. As such, we recommend its inclusion in new QSAR/QSPR projects. Overall, our findings call for further in-depth empirical and theoretical analyses, to increase the practical relevance of ML models for drug development and chemoinformatics.

\section*{Supporting Information}
The following file is available free of charge: \textit{supplementary.pdf},
containing statistics of the benchmark and external validation datasets,
including the curation of the latter; the evaluated model variants and the
selected representative of each architecture; classifier hyperparameter grids;
details of the CLAMP retraining; hardware details; full Bayesian Bradley-Terry
statistics, with decisions for different ROPE intervals and under the strong
interpretation; the sensitivity of the hierarchical t-test to the prior on
$\nu$; mean ranks and AUROC of all model variants and classification heads,
and the number of wins and near-wins of all models; and computation times of
all models.

\section*{Acknowledgements}

We thank the BIT Student Scientific Group for computational resources and assistance for this project. We thank Michał Lampert for initial experiments and analyses. We further thank Aleksandra Elbakyan for her work and support for accessibility of science.

\section*{Data and Software Availability}

All datasets and benchmarks used are publicly available at \url{https://github.com/MLCIL/benchmarking_molecular_models}. Datasets from MoleculeNet \cite{MoleculeNet} and TDC \cite{TDC} benchmarks are also available on their respective websites. External validation datasets are available in the GitHub repository.

\section*{Author contributions statement}

Contributions follow the CRediT format. M.P. was responsible for software, validation, data curation, and writing. J.A. was responsible for conceptualization, methodology, investigation, data curation, and writing. W.C. was responsible for writing, review, and supervision.

\section*{Competing interests}

Jakub Adamczyk and Mateusz Praski work as chemoinformaticians, but their commercial work has no relation to this manuscript. The authors declare no competing financial interest between their scientific and commercial work.

\section*{Funding}

This work was supported by general funding from the Ministry of Science and Higher Education to AGH University in Krakow, and by project Excellence Initiative – Research University (IDUB) for AGH University of Krakow.

\printbibliography

\newpage

\noindent
\begin{minipage}[b][1.75in]{3.25in}
  \centering
  \includegraphics[width = 3.25in, height = 1.75in, keepaspectratio]{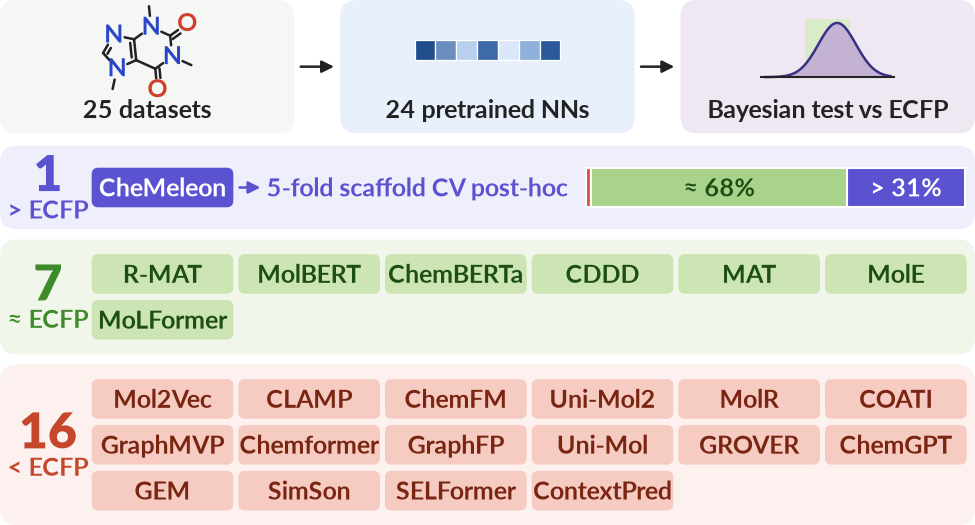}
\end{minipage}

\end{document}